# Implementation and Evaluation of a System for Assessment of The Quality of Long-Term Management of Patients at a Geriatric Hospital

Erez Shalom, Ayelet Goldstein, Roni Wais, Maya Slivanova, Nogah Melamed Cohen and Yuval Shahar

**Background**

The rapid growth of the aging population over the past decades has become an important challenge, capturing the attention of the world. Furthermore, in recent years, there has been a growing shortage of professional care, leading to significant therapeutic burden. The use of a clinical decision support system for assessing the quality of care, based on computerized clinical guidelines (GLs), is likely to improve medical care, reduce costs, save time, and enhance the clinical staff's capabilities.

**Objectives**

Implement and evaluate a computerized system for assessment of the quality of the care, in the domain of management of pressure ulcers of geriatric patients admitted to a Geriatric Medical Center. Assess the quality of care, by investigating the level of compliance of the clinical staff to the GLs over significant time periods, using retrospective longitudinal data.

**Methods**

We designed and implemented a computational quality-assessment system, based on a fuzzy temporal logic-based algorithm, which allows for partial compliance of the clinical staff to quality metrics (i.e., assigning a score between 0 and 1 to each action, or predefined group of actions, based on their value and on their temporal aspects).

We represented the Clinical GL for the treatment of pressure ulcers in a computerized manner, in collaboration with the clinical staff. We also obtained longitudinal data for 100 random patients from the local EMR system.

We then performed a technical evaluation of the system, checking its applicability and usability, followed by a functional evaluation of the system. The functional evaluation was performed by a highly experienced nurse who was not part of the knowledge-acquisition team. The evaluation investigated the quality metrics given to the compliance of the medical's staff to the protocol. We compared (1) the scores given by the nurse when supported by the system, to (2) the scores given by the nurse without the system's support, and to (3) the scores given by the automated system. We also measured the time taken to perform the assessment with and without the system's support.

**Results**

There were no significant differences ($P < 0.05$) in the scores of most measures given by the nurse using the system, compared to the scores given by the automated system. There were also no significant differences ($P < 0.05$) across the values of most quality measures given by the nurse without the system's support compared to the values given by the nurse with support of the system. Using the system, however, significantly reduced the average assessment time taken by the nurse to score each patient across all quality measures.

**Conclusions and future work**

The automated quality-assessment system provides, for complex clinical care over significant time periods, quality-measure scores that are very similar to that of an experienced senior nurse, showing that using an automated quality-assessment system, such as the one we have developed, may enable a senior nurse, after a short training, to quickly and accurately assess the quality of care. In addition to its accuracy, the system considerably reduces the time taken to assess the various quality measures; time saving is expected to increase as users become more experienced operating the system. Decision-support systems such as the one we have developed and assessed may empower the nursing staff, enabling them to manage more patients, and in a more accurate and consistent fashion, while reducing costs.

## 1. Background

### 1.1 Population aging challenges

For the past decades, following baby boom events and improvement in life expectancy, the world has been experiencing a gradual increase in the proportion of citizens aged over 65. According to the UN's projection, the proportion of the world's population aged over 65 will rise from 10% in 2022 to 16% by 2050 [1].

An ageing population presents many challenges for health systems, including a shift towards care-based and end-of-life services, which leads to an increase in medical costs and long-term care [2]. On the other hand, there is a growing shortage of professional care, leading to a heavy therapeutic burden on the care staff, as there are fewer doctors, interns and nurses in the nursing system. To improve medical care for the aging population and curb its costs, many efforts are being made to base it on up-to-date *Clinical Guidelines* (GLs).

According to the IOM's report "To Err is Human" from 2000, close to 100,000 patients die each year as a result of medical errors [3]. A report from the U.S. Department of Health for 2010 indicates that medical errors occur in 13.5% of hospitalizations, with a cost of $4.4 billion for the US' health insurance company, and that 44% of these errors were defined as preventable [4].

The use of GLs may reduce the number of adverse events, such as misdiagnosis of a drug, or the occurrence of a preventable infection [3], and reduce the variability of care: over the past two decades, studies have been published showing that GLs improve treatment uniformity, which may both increase the quality of physicians' decisions and increase patient survival rates while reducing morbidity, and may even reduce treatment costs [5-6]. The latest IOM report recommends using GLs to reduce the rate of preventable adverse events [7].

However, despite all the enormous benefits inherent in the use of GLs, its usage encounters a number of barriers: Most GLs are represented in free text and are usually inaccessible to the doctor at the point of care, especially when the doctor usually does not have the appropriate time and tools to search for the most apropriate GLs for the current patient. These reasons lead to low responsiveness and adherence to GLs on the part of physicians and low percentages of adoption among the medical staff [8].

Thus, there is a need to develop a *Clinical Decision Support System* (CDSS) that will assist the physician in automatic application of GLs at the point of care and in automated assessment of the quality of that application by the clinical staff. We had previously focused on the first aspect, which we shall now briefly describe.

### 1.2 The Picard System

In recent years, we have developed a CDSS that assists care providers of various types in the automated application of GLs at the point of care. The system allows acquisition and formal representation of GLs in a formal manner comprehensible and interpretable by a computer, given the longitudinal digital patient record. We implemented a GL application engine called Picard [9]. We evaluated the Picard GL-application engine in a number of studies [10-12] that demonstrated conclusively that using CDSSs increases physicians' compliance to the GL (*completeness*) and the percentage of the physicians' actions that are justifiable by the GL (*correctness*) and reduces costly redundant expenditures, such as unnecessary laboratory tests.

As part of our main research, we developed a system for application of a GL for the treatment of long-term preeclampsia in the Department of Obstetrics and Gynecology at Soroka University Medical Center, one of the biggest medical centers in Israel. The guideline for treating preeclampsia was represented on a knowledge base used by the Picard System. We simulated medical records to fit six different disease scenarios, which included a total of 60 decision points. 36 obstetricians (24 residents and 12 board-certified experts) participated in the experiment. Each obstetrician made his decision in half of the instances of the six scenarios without any computer assistance, and in the other half, after being provided with the CDSS's recommendation.

The evaluation metrics included **correctness** of the actions (i.e.: what percentage of the clinician's decisions were correct, according to the guideline) and their **completeness** (i.e.: what percentage of the guideline's recommendations were indeed performed by the clinician) compared to the treatment based on the guideline. The subjects also filled out evaluation questionnaires regarding the use of the system. We found a significant improvement ($P < 0.05$), while reducing variance, in the **completeness** of the decisions: from a mean of 41% completeness without the use of Picard, to a mean of 93% with the use of Picard; we also found a significant improvement, while reducing variability, in the **correctness** of the decisions: From a mean of 27% correctness (and necessity) without the use of Picard, to a mean of 91% correct and necessary actions out of the clinicians' actions, with Picard's assistance. An additional 68% of the actions performed without CDSS help were not an error, but were redundant; this percentage fell to 3% with Picard's assistance. The improvement in the evaluation metrics and the reduction of the variance were observed throughout all scenarios, for all subjects, in all specialization's levels and for all decisions' type, thus considerably reducing what Kahneman et al. refer to as decision *noise*, which depends on the actual decision maker, on the context of the decision, or on the interaction between the decision maker and the case at hand [13]. Subjects also evaluated positively the potential use of the system. This study showed that using a CDSS could prevent the dependence of the quality of the decision on the specific physician, the clinical scenario, or the nature of each particular decision. This will ensure a better decision-making process, and a higher compliance to GLs.

In another, much larger-scale (geographically) experiment, the Picard System was a key component in the European Union's 7th Research and Development Project (7FP), the MobiGuide project [14]. This study used the patient's smart phone and a group of CDSS servers, to monitor the patient, using sensors located on the patient's body or at their home. Chronic patients were monitored by ambulatory clinics belonging to two hospitals: in Barcelona, Spain, at the Sabadell Hospital, for the treatment of patients with gestational diabetes; and in Pavia, Italy, at the Fondazione Salvatore Maugeri Hospital, for the treatment of patients with atrial fibrillation. The Picard system sent alerts when necessary to the patients or to their physicians, and provided customized, context-sensitive and patient-specific recommendations, based on the appropriate medical guideline, to the patients' mobile phone or to the care provider's desktop computer. As a result, the treatment was tailored to the patient, and varied if necessary according to the patient's clinical, physical, mental and socioeconomic status, and according to the patient's preferences regarding the treatment and their physical environment. The result was a high compliance by patients and care providers to the system's recommendations, and several indications of improved process of care and of its clinical outcomes[15].

### 1.3 The need for computer-based quality assessment

*Health Maintenance Organizations* (HMOs) are constantly accrediting their hospitals, requiring them to comply with standards of quality and safety of care and the implementation of at least five GLs per year. Accurate and timely assessment of the quality of care and of the compliance to establish, evidence-based GLs, might support the effort to continuously enhance the quality of care and reduce its variance.

An example for one of the most common guidelines in the geriatric wards, on which we focused in the current study, is the treatment of pressure ulcers (bedsores), a problem from which 30% of hospitalized patients suffer, and about 17% to 28% in prolonged hospitalization [16]. The treatment cost is high and in the United States annual expenses due to pressure ulcers reach $9.1 to $11.6 billion[17]. In addition, inpatients have other risk factors that should be monitored and checked in an organized manner as part of additional guidelines, to identify deterioration of the patient's condition, such as restriction of mobility, malnutrition, lack of control of the sphincters and cognitive impairment.

The Herzfeld medical center in which we performed the evaluation of the quality assessment CDSS that we shall describe belongs to the largest HMO in Israel (which is one of the largest in the world, since it manages close to 60% of the Israeli population). It has six inpatient wards (about 40 beds in each ward) and a dialysis institute (about 68 beds), with an average hospital stay lasting between weeks and several months, a fact that increases the risk of pressure sores in inpatients. The pressure ulcer treatment is performed by nurses who need to remember the rather complex textual guideline, making the adoption and implementation of the guideline by the already overloaded nursing staff quite difficult. As a result, the guideline's compliance is affected, leading to a decrease in the quality of care and even to medical errors that could have been avoided.

Thus, a GL-based CDSS that can monitor retrospectively and accurately the level of performance of each GL and of GL component, in crucial and complex GLs such as the GL for pressure ulcers, may improve medical care management, reduce its variability, and might expand the capabilities of the nursing staff and support the many tasks assigned to them. In the current study, we have designed and implemented such a CDSS and evaluted its feasibility within the Herzfeld griatric center.

## 2. Methods

### 2.1 The DeGeL Digital Guideline Library framework

In this research, we employed the *Digital Guideline Library* (DegeL) architecture [18] previously developed at the BGU Medical Informatics Research Center. DeGeL includes tools for converting GLs from its textual representation to a formal, machine comprehensive representation in the Asbru GL specification language [19-20]. This process is called **knowledge acquisition**. The uniqueness of the DEGEL system is in the gradual conversion process of the GL from its textual, *unstructured* (free-text) representation (which is usually done by medical experts), through its *semi-structured* representation (which is usually performed by knowledge experts and medical experts), to its *formal, structured* representation (usually prformed by knowledge experts).This conversion is done using graphical tools, for acquiring knowledge, such as "Gesher", developed in our lab. (see Figure 1). These tools have been evaluated in a series of studies [21-23], which had demonstrated that physicians are able to use the tools and to perform the construction process of the GLs in a complete way, with the knowledge engineer assisting in its formal representation if necessary. The formal representation was saved in a knowledge base.

The guideline's knowledge includes:
(a) *Procedural knowledge* ("How to") (e.g., which medications to administer, how much, and when, and under what conditions to start or stop them, and from which plans is the GL composed, including sequential, parallel, and periodic plans). The procedural knowledge of GLs in our studies was usually represented using the *Asbru* GL-specification language [24];
(b) *Declarative knowledge* ("What is") (e.g., what does "renal insufficiency" or "bone-marrow toxicity grade 2" or "deteriorating liver functions over the past year" mean, what is its precise definition, which logical and probabilistic components is it composed from [e.g., does "moderate anemia" on Monday and Thursday mean that the patient had a week of "moderate anemia"?). Declarative knowledge in our studies was usually represented using the *Temporal-Abstraction Knowledge* (TAK) language, an extension of the *Knowledge-Based Temporal Abstraction* (KBTA) ontology [25] and the CAPSUL periodic temporal patterns language [26].

(c) *Quality-Assessment knowledge* ("Why, or to achieve What"?) (e.g., what is the intended patient *outcome* of each part of the GL or of the overall GL? What is the intention underlying a correct *process*, such as "reduce the blood pressure *by using diuretics or beta-blockers*"? How do we score the quality of a monitoring process in which the nurse visited the patient on average twice out of the intended three times per day?). Quality-assessment knowledge was represented in our study using the TAK language with a fuzzy temporal logic extension. We shall delve more deeply into this issue when we discuss our quality-assessment methods.

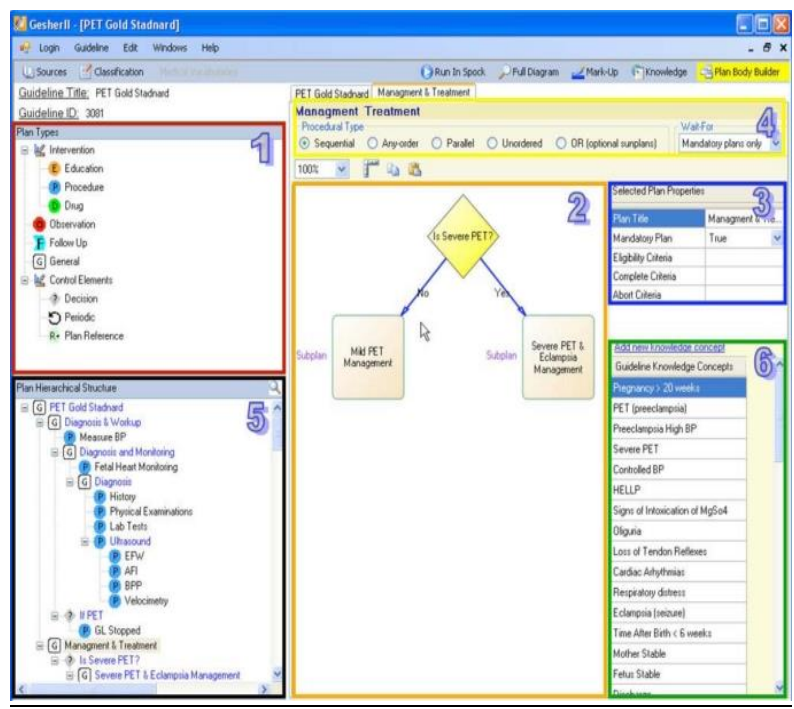

**Figure 1.** The interface of the GESHER tool, which assists in the performance of knowledge acquisition. Using Gesher, clinicians can define, with the help of knowledge engineers, the hierarchy of treatment plans, i.e., the procedural knowledge of the GL. Here, we see part of the plan "Management and Treatment" of the preeclampsia-management GL. The overall plan starts by an assessment of the patient's condition, classifying it as either a severe or a moderate preeclampsia. The user selects management plans of different types (Frame 1) and adds them to the hierarchical flow chart (Frame 2). For each plan, several properties can be defined (Frame 3). For plans composed of several sub-plans, procedural knowledge attributes, such as the order of performance of the plans (parallel, serial, etc.) are also displayed (Frame 4). The sub-plan hierarchy is displayed as a Tree view (Frame 5). At this stage, the expert also defines a list of declarative knowledge concepts relevant to the GL, such as "Severe Hypertension", which will be elaborated in the next stages of knowledge acquisition and can be selected for further use, as needed (Frame 6).

For the purpose of knowledge acquisition, together with the medical staff, which included a nurse in charge of quality at the hospital and the head of the department, we defined the pressure ulcer GL in a formal representation. The GL was based on an existing Clinical Guideline for the prevention and treatment of pressure ulcers for hospitalized patients, and included 4 main stages: (1) Admission, (2) follow-up, (3) follow-up and prevention, (4) follow-up, prevention and treatment. Each stage has a different pre-condition, and includes several procedures such as bandages that should be applied, and different operations that need to be performed. The first step towards building a computerized GL, was to perform a Ontology Specific Consensus (OSC), shown in Figure 2, validated with the clinical staff in order to ensure it represented the actual clinical practice. Then we continue to structure the GL, leading to the final representation of the computerized GL. Figure 3 shows the procedural knowledge of the GL modeled in the "Gesher" medical knowledge acquisition tool, and its various stages.

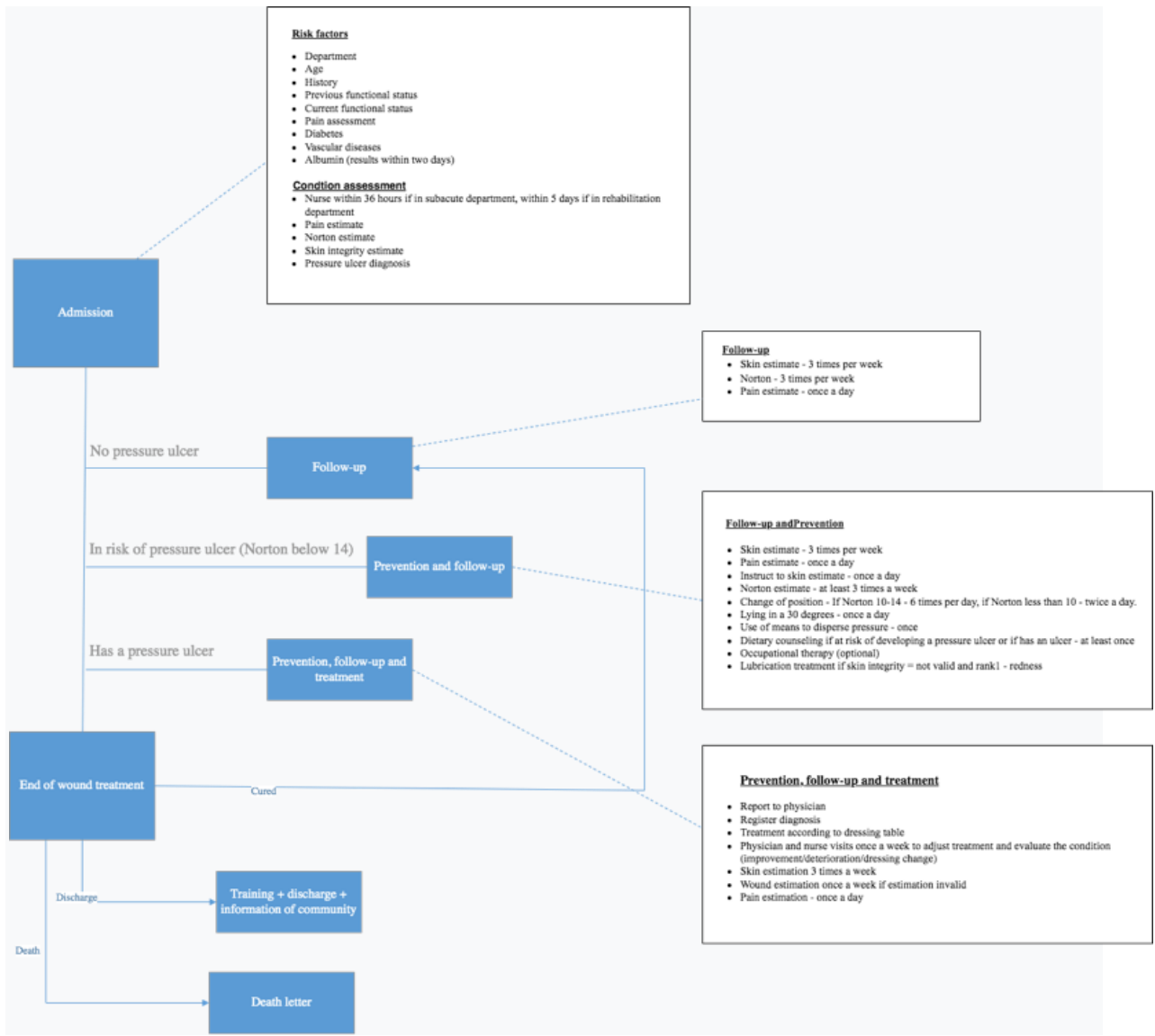

**Figure 2.** The protocol for the treatment of pressure ulcers in its textual form, as defined by the domain experts. The protocol has four main steps: (1) Admission, (2) follow-up, (3) follow-up and prevention, and (4) prevention, follow-up and treatment.

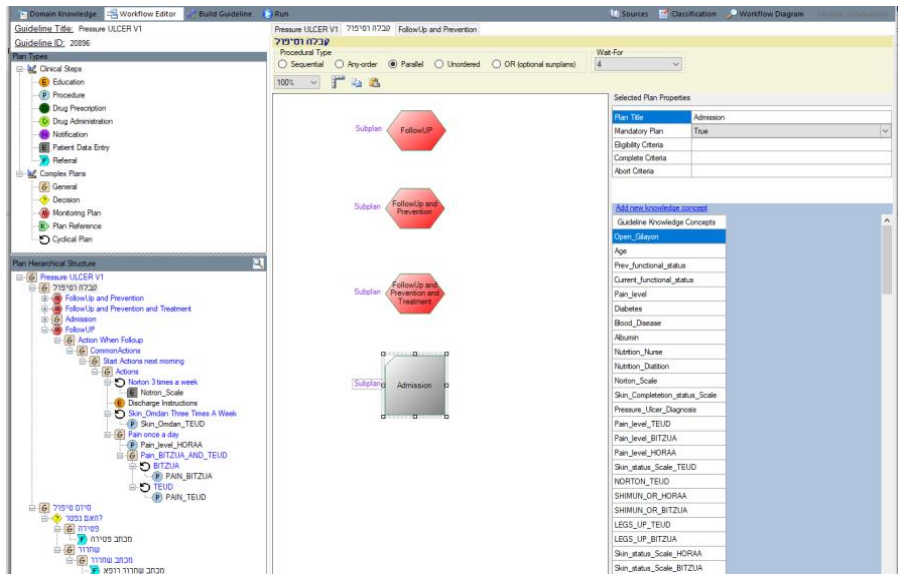

**Figure 3.** The pressure ulcer protocol formalized within the Gesher medical knowledge acquisition tool. The plan hierarchy is displayed on the bottom left; the declarative concepts are shown on the bottom right.

## 2.2 Constraint types

In order to calculate the compliance of the actions described in the protocol, we defined several categories of constraints for an action :

1. **Binary constraint** - actions that need to be performed only once, e.g performing the Norton test.
2. **Cyclical (periodic) constraint** – actions that are performed in a pre-defined frequency, e.g, estimating pain once a day.
3. **Time constraint** - actions that need to be performed within a certain period of time from a referenced event ,e.g. an albumin test that must be performed within two days after receiving the patient.
4. **Entry-condition constraint** - actions that need to be performed when a specific *entry condition* is met, e.g. a patient at specific ward should receive nutritional advice from a dietitian.
5. **Order constraint** - actions that need to be performed in a specific order (i.e., before or after other action), e.g. taking pain killer need to performed only after a pain assessment test.
6. **Multiple constraints** - actions that need to be performed when at least one out of several conditions are met, e.g., given a symptom of redness in the heels, a silicone bandage or an olive oil bandage can be performed.
7. **Combination constraints** - A combination of constraints from the above (1-6), for example, the action "estimating pain once a day in the follow-up phase", have a *cyclical* constraint (need to be performed once a day), an *order* constraint (need to be performed after an instruction to perform this action) and a *binary* constraint (the action needs to be indeed performed)

The constraint category defines the calculation type that will be used to assess the action's compliance score.

## 2.3 Calculation Types

The action's score is calculated in one of the following ways, according to its constraint type:

1. **A Binary calculation:** The score will be either 1 or 0, according to whether the action was performed or not, without penalty for unnecessary actions.
   This calculation is used for actions with a *binary* constraint, stating if the actions was performed or not. For example, the action "Norton test performed once at the admission stage will receive either 1, if performed or 0, otherwise.

2. **A proportional calculation:** The score is calculated according to the number of times (cardinality) that the action was performed, out of the total times that the action should have been performed. In general, The action's score is calculated according to the proportion of the time frame in which the action was actually performed, relative to the time frame required in the protocol.

   This calculation is often used for actions with an entry-condition constraint – calculating, out of all the times when the entry condition has met, how many times the action was actually performed. For example, out of the times that the entry condition "pressure ulcer color is red with no slops or little slops" was met (the denominator), how many times the correct bandage was applied (the numerator).

   A proportional calculation is used also for actions with an *order* constraint – calculating out of all the times the action was performed, how many times it was performed in the correct order. For example, out of the times an instruction to perform a pain test was given (the denominator), how many times the paint test was performed (the numerator).

   A proportional calculation is also used for actions with *multiple-dependencies* constraint – calculating out of all the times when at least one of the multiple entry conditions was met, how many times the action was performed.

   A proportional calculation is also used for actions with a *cyclical* constraint – calculating the performance rate in the expected time frame. The score is calculated using the formula:

   $$\frac{cardinality\_Performed}{cardinality\_Expected * Time\_Proportion}$$

   Where *cardinality_performed* is the number of times the action was actually performed, *cardinality_expected* is the number of times the action was expected to be performed, and *time_proportion* is the time interval between each instance, using a uniform distribution. For example, the action "perform a Norton test three times a week" has a time-window of one week, in this case, the interval between each instance (*time_proportion*) is expected to be 1.75. If four days has passed since the beginning of the week and a Norton test was performed only once, the score will be $\frac{1}{3*\frac{4}{7}} = 0.5833$

   In the case of partial intervals observed, the cardinality expected is adjusted according to the proportion of the observed interval. For example, for an action expected to be performed 3 times a week, for an observed partial interval of 3 days since the proportion of the observed interval is 3/7 , the cardinality expected will be adjusted from 3 to $\frac{3}{7} * 3 = 1.28$

In case of two consecutive intervals, we measure the gap between the last measurement in the first interval to the first measurement in the second interval.

3. **A [Temporal] Fuzzy logic calculation**: – A [temporal] fuzzy logic function allows for a partial scoring of adherence to the protocol's recommended action, using a fuzzy logic algorithm based on a trapezoid schema: Each trapezoid is defined by up to 4 points. The score is determined by the trapezoid's points as shown in Figure 4. The X-axis describes the time interval between an action and a subsequent action, and the Y-axis describes the score [0..100] given for that action given the interval. There are actions for which vertex A and B have the same value, and their X-axis value is even 0, i.e. no minimum interval is required between operations, and there are actions whose fuzzy-logic description requires all four vertices.

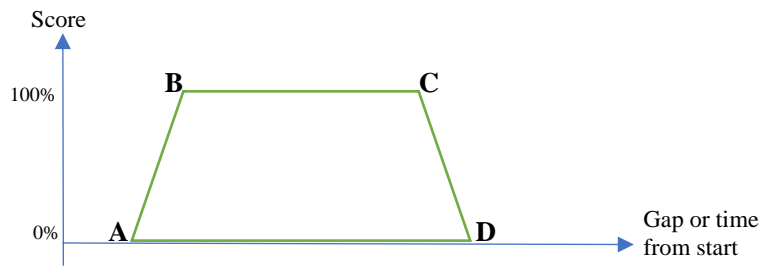

**Figure 4 .** The trapezoid maps the allowed time frame of an action to a score in the range [0-1]. Point A defines the value below which the performance is not valid, and therefore the score will be 0. The interval of values from Point B to Point C defines the range of values for which the performance is complete. Any value between A and B is a partial performance, and as it gets closer to B the score will get closer to 1. Point C defines the value from which the performance is partial, and therefore the score is partial. Point D defines the value beyond which the performance is not valid, and a therefore the score of 0. Any value between C and D (like any value between A and B) denotes a partial performance, and thus is assigned a partial score.

A fuzzy-logic calculation is used for actions with a time constraint – calculating the score according to the time defined in the trapezoid, from the beginning of the entry condition(s) until the action is performed.

For example: for "perform an albumin test during reception within a day", the protocol defines that if the test returns up to 72 hours from the admission, the score maximal, will be 1. If test is done within 72-120 hours a partial score will be assigned (using a linear function, between the points (72,1), (120,0). If test is done in more than 120 hours the score will be 0. The trapezoid in this case is shown in Figure 5.

Thus, for example, if the albumin test was performed after 80 hours from admission, the score will be 0.833:

$m = \frac{0-1}{120-72} = -\frac{1}{48}, \quad y = -\frac{1}{48}x + b$

$for\ (x = 72, y = 1):\ b = \frac{1}{48} * 72 + 1 = \frac{5}{2}$

Score= $= -\frac{1}{48} * 80 + 2.5 = 0.833$

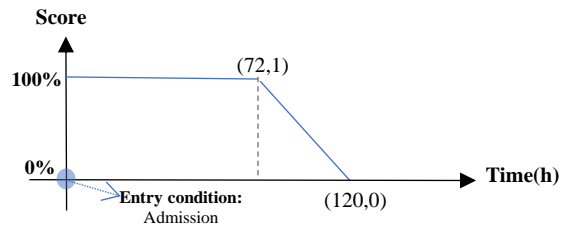

**Figure 5.** A fuzzy-logic Trapezoid representing the time constraints of the action "the test should be performed within 72 hours from admission, and not latter than 120 hours". The trapezoid defines that a maximal score of 1 if the test is done within 72 hours, a score of 0 if the test is done after 120 hours and a partial score between 0 to 1 if test is done between 72 to 120 hours, using a linear function.

A fuzzy-logic calculation is also used for actions with a *cyclical constraint* – calculating the score according to the gap between each instance of the action.

For example: for "perform a Norton test three times a week", the protocol defines that the gap between the instances will be assigned a score of 1, if it is less than 72 hours, a partial score between 1 and 0 if the gap is between 72 and 80 hours, and a score of 0 if the gap is greater than 80 hours. Therefore, a test with a gap of 75 hours between the instances will receive, using the default linear function, a score of 0.625, as shown in the trapezoid displayed in Figure 6.

$$m = \frac{0-1}{80-72} = -\frac{1}{8} \quad , \quad b = \frac{1}{8} * 72 + 1 = 10 \quad ,$$

$$\text{score} = -\frac{1}{8} * 75 + 10 = 0.625$$

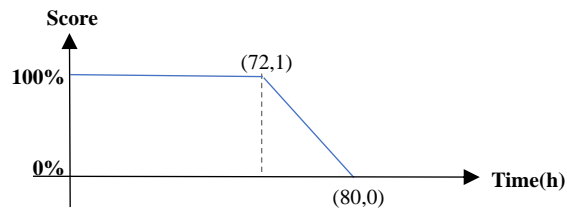

**Figure 6.** A fuzzy logic trapezoid representing a cyclical constraint, in which a full score is assigned to the actions if the gap between instances is smaller than 72 hours, a partial score is received if the gap is between 72 and 80 hours, and a score of 0 is assigned if the gap is greater than 80 hours.

### 2.4 Action weights

In addition to the quality-assessment score, each action might also have a weight representing its importance in the final calculation of the GL's adherence score, if needed. The final calculation of the GL's overall adherence score is achieved by multiplying the action's score by the action's weight and summing the result over all actions. The system enables the knowledge engineer to specify weights to each action within a step, and also to each step within the protocol. This capability enables the knowledge engineer to penalize non-performance of actions that are considered to have greater importance, thus emphasizing the critical hospitalization stages.

In addition, the system allows the knowledge engineer to assign weights to the action's different components. For example, different weights can be assigned to the explicit request (instruction) for performing an action's, and to the actual performance of the action. Different weights can also be given for an action's performance frequency and to the correct order of it performance. The final score of the action will be calculated according to the weighted average of the action's components.

A description of the weights of each step and action acquired from the domain experts appears in Appendix A. Table 1 shows an example of the follow-up stage.

**Table 1.** Weights acquired for the follow-up stage. In addition to the weight of the stage (22%), weights are given also for each action within the stage. For each action, weights can be given for parts of the action (e.g., the temporal order and the actual performance of the action) and also for different components of the score (e.g. frequency and order).

| Stage | Stage weight | Component | Weight | Component | Weight | Component | Weight |
|---|---|---|---|---|---|---|---|
| Follow up | 22% | Pain-once-a-day | 30% | Performance | 0.5 | Frequency | 0.5 |
|  |  |  |  |  |  | Order | 0.5 |
|  |  |  |  | Command | 0.5 |  |  |
|  |  | Skin-3-times-a-week | 35% |  |  |  |  |
|  |  | Norton-3-times-a-week | 35% |  |  |  |  |

### 2.4 The Quality-Assessment Business Intelligence (BI) dashboard interface

We developed a *Business-Intelligence* (BI) dashboard interface that enables the medical staff to see the scores of the actions in different protocol's stages. Scores can be displayed for a selected group of patients, time frames, and specific wards, in different levels of granularities (e.g. general score of the follow-up phase, and drill-down into the sub-scores of each action within the stage).

The score of a specific time frame is the weighted average of the scores of all the sub-intervals within that time frame, taking into account the interval's sizes. The score of a time frame allow us to compare between the medical staff's responsiveness during different time frames. For example, differences in the compliance during weekends compared to midweek, night shifts versus day shifts, etc.

Scores can also be calculated for a specific patient or group of patients. The score of a group of patients is the average score of the patients within that group.

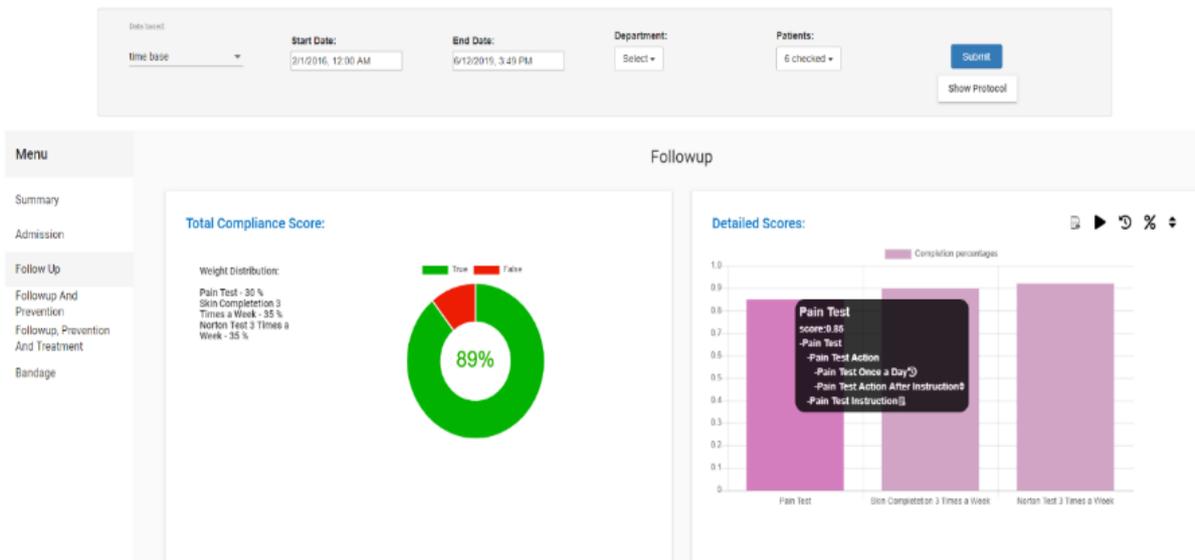

**Figure** 7. The BI Dashboard. On the left are displayed the different stages of the protocol: "admission", "follow-up", "follow-up and Prevention", "FollowUp, Prevention and Treatment" and "bandage". The score of the current stage (follow-up), 89%, is calculated based on the weights and scores of each action this stage: (1) pain test, weighted 30% with a score of 85, (2) "skin completion 3 times a week" weighted 35% with a score of 90 and (2) "norton test 3 times a week" weighted 35% with a score of 92.

In the ruler at the top of the dashboard, we can select specific time-frame to focus on, as well as specific departments or group of patients.

## 4. Evaluation

### 4.1 The data

After getting the *institutional review board* (IRB) approval at the Herzfeld geriatric center to access retrospective data, we retrieved 100 random longitudinal patient records from the HER in which the term "Pressure Ulcer" appeared anywhere in the problems list. Each record included longitudinal data of several months of patient hospitalization in one of the departments participating in the study from November 2016 to November 2017. The data mainly included measurements related to pressure ulcer treatment such as: age, Norton scale estimate, degree of ulcer severity, color, size and depth of ulcer, secretion and odor, relevant background diseases, nutritional estimate, dietary advice, skin integrity, posture change, pressure-dispersing measures, type of dressing, pain estimate, MUST estimate and patient guidance.

### 4.2 The Knowledge

During the formalization process, we acquired the declarative knowledge, which included more than 140 different knowledge concepts that have been mapped to the hospital's medical database. Table 2 describes the distribution of knowledge concepts acquired, using the TAK language (see Section 2.1) in the different treatment stages.

**Table 1**. Knowledge concepts acquired

| Stage | Number of TAK files |
|---|---|
| Admission | 12 |
| Follow-up | 7 |
| Follow-up and prevention | 43 |
| Follow-up, prevention and treatment | 26 |
| Bandages | 52 |
| Total | 140 |

### 4.3 The Technical evaluation

In the technical evaluation, using retrospective patient longitudinal data, we evaluated the feasibility of the system for quality assessment of the acquired GL.

We performed the technical evaluation in two steps: In the first step, we evaluated the algorithm, assessing whether the resulting output was as expected. In the second step, we evaluated a list of pre-defined quality-assessment patterns such as "performance of the action before the instruction for it was given", assessing their accuracy. For this purpose, we simulated data that were sufficiently realistic as to have a structure similar to that of the medical records we had retrieved, and used a sophisticated visualization tool that displays in the data time-dependent templates for individual and multiple patients (VISITORS [27]) to assess the correctness of each pattern, regarding time and value.

In both steps, all of the technical tests of the computational system were successful.

The BI interface's usability was evaluated by the senior nurse, using the standard usability questionnaire SUS [28], with a score of 90, indicating a high level of usability. The details of the questionnaire and of the results appear in Appendix B.

Thus, we felt sufficiently confident to move on to the functional evaluation.

### 4.4 The Functional evaluation

For the functional evaluation, we selected 29 random patients and 30 quality metrics of care, relevant to the GL, selected with the help of our domain experts. A highly experienced senior geriatric nurse who was not part of the research team calculated the quality assessment scores of the treatment for all metrics, for each patient. Half of the patients were assessed by the nurse using the new BI interface, after a short training with the system, and half of the patients were assessed manually, without the interface, using only the raw patient data.

We compared the scores assigned manually by the nurse to each quality metric of care, to the score given by the system, using a number of statistical tests such as: Spearman's non parametric correlation (to compare the ranking of the scores, for each metric, across all patients), and Pearson's Correlation (to compare the linear ordering of score values).

We also measured and compared the average time taken by nurse to perform the evaluation without and with the system for each group of patients.

Below we present the evaluation results.

## 4.5 The Functional Evaluation results

### 4.5.1 Comparison of Manual QA to Automated System QA

In the manual evaluation, the senior geriatric nurse assigned scores manually to 15 out of the 29 randomly selected patients, for several quality metrics selected at random for each patient (a range of 9 to 17 metrics, with an average of 13 metrics per patient). Using an Excel spreadsheet as an electronic medical record, the nurse could relatively easily find the patient data needed to determine each quality metric's score.

The results, comparing the score assigned manually by the nurse to the score calculated by the automated system, are shown in Table 3. Note that a single patient can have several occurrences of the same quality metric but at different stages of the protocol. For example, for the quality metric "reference to a dietician consultation" appearing 10 times in the protocol the spearman test comparing the ranks of the scores given by the system to the ranks of the scores given by the nurse had a result of R=0.853, p<0.01.

**Table 2.** Comparison between the quality metrics' scores assigned manually by the nurse compared to the scores calculated by the automated system.

|  |  | Spearman test | | Pearson test | |
|---|---|---|---|---|---|
| **Quality metric** | **#Appearances** | **R** | **p-value** | **P** | **p-value** |
| Referral for dietary advice | 10 | 0.913 | 0.00022 | 0.981 | 0.0000005 |
| Nutrition (must) | 10 | same |  | 0 | 1 |
| skin integrity estimation - once a week | 11 | 1 |  | 1 |  |
| Pain estimation - once a day | 14 | 0.596 | 0.02449 | 0.848 | 0.00012 |
| Norton estimation - admission | 15 | same |  | 0 | 1 |
| Skin state estimation - 3 times a week | 15 | 0.559 | 0.03028 | 0.33 | 0.22966 |
| Skin state estimation – on admission | 15 | 0.853 | 0.00005 | 0.853 | 0.000052 |
| Change of position - according to frequency | 15 | -0.034 | 0.90425 | -0.03 | 0.91547 |
| Instruction Pain assessment - once a day | 15 | same |  | 0 | 1 |
| Instruction for change of position | 15 | -0.16 | 0.56894 | -0.155 | 0.581232 |
| Document Norton - 3 times a week | 15 | 0.548 | 0.03444 | 0.834 | 0.000111 |
| Execution of dietary advice | 21 | 0.684 | 0.00062 | 0.734 | 0.0001521 |

It can be seen that most of quality metrics are significantly similar between the two vectors (nurse scores and system scores, for each metric, across all patients), some of them being simply identical.

Note in Table 3 that the correlation in the metric "change of position" (both the instruction and its performance) is not high. It turned out on further inquiry that the reason was because in rehabilitation wards, patients were allowed to get out of bed in the morning and in the evening; and since it did not matter *when* that was performed, the change of posture was simply not recorded in the medical record of the patient. It also turned out that this practice is "out of the protocol" and was done "unofficially" by some nurses in certain wards. It is worth noting that the fact that this informal practice was discovered by comparing between the scores of an automatic quality control system to the scores of a senior nurse is interesting in itself.

**Thus, in practice, the system seems to assign quality assessment scores that are very similar to that of a highly experienced senior geriatric nurse.**

### 4.5.2 Manual Quality Assessment Performance time

Figure 8 displays the time taken by the highly experienced senior nurse to manually calculate the scores of all metrics using the prepared spreadsheet for each patient.

As can be seen, the time for performing the quality assessment varies among patients, with an average of 1,039.29 ± 242.4 seconds (i.e., a mean of approximately 17.3 minutes) to score on average 13 quality metrics per patient. Thus, scoring the QA scores for all of the 100 patients would be expected to require approximately 1730 minutes, or 28.8 hours, even for only the 13 metrics, on average, selected. It should be noted that using the hospital's EMR would be likely to have required an even longer time, due to the need to search for each relevant piece of raw data.

The automated system's calculation was instantaneous.

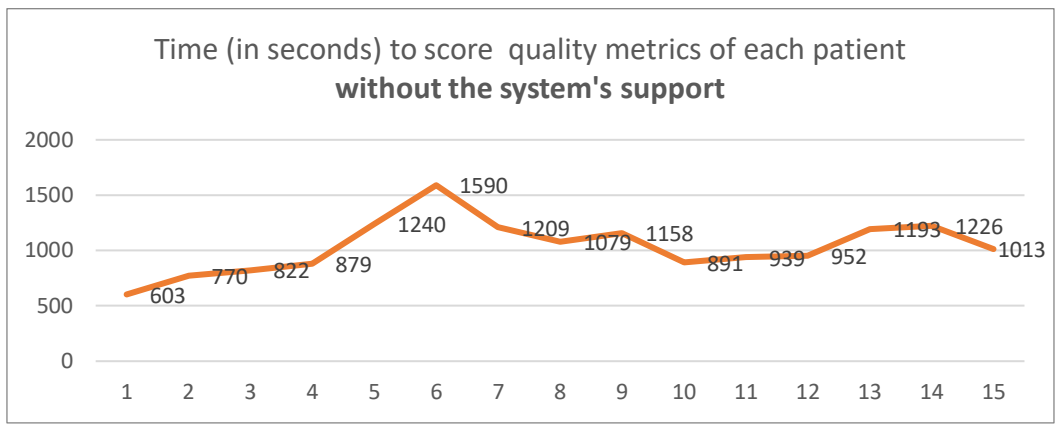

**Figure 8.** Time taken by the nurse to manually score all the quality metrics of each patient without the system's support.

### 4.5.3 Comparison of System-supported Manual QA to Fully Automated System QA

Using the QA system's support, the same senior geriatric nurse using the BI interface, assigned scores to the quality metrics to the other 14 patients out of the 29 randomly selected ones (a range of 9 to 15 metrics, with an average of 12 QA indicators per patient). For each metric, the nurse searched for its value, using the interface, and wrote separately its score for each specific patient. Note that the nurse was new to the system, so some errors were to be expected.

We compared the scores assigned by the nurse with the system's support to the scores given by the automated system and retrieved by one of our research team's medical knowledge engineers, who was an experienced user of the system. Table 4 displays these results.

**Table 3.** Comparison of the quality metrics' scores assigned by the nurse when using the system's support, to the scores assigned by the system and retrieved by an expert user.

|  |  | Spearman test | | Pearson test | |
|---|---|---|---|---|---|
| **Quality metric** | **#Appearances** | **R** | **p-value** | **P** | **p-value** |
| Norton estimation - admission | 14 | 1 | same | 1 | same |
| Skin state estimation - 3 times a week | 14 | -0.11289 | 0.70079 | -0.078566 | 0.78948 |
| Skin state estimation – on admission | 14 | 1 |  | 1 |  |
| Skin integrity estimation - once a week | 12 | 1 | same | 1 | same |
| Pain estimation - once a day | 13 | 0.72083 | 0.0054 | 0.89221 | 0.00004121 |
| Performance of correct bandage for "red without or little secretion " | 13 | 0.88396 | 0.00006 | 0.87344 | 0.00009 |
| Performance of dietary advice | 15 | 1 | same | 1 | same |
| Instruction Pain assessment - once a day | 14 | 0.746 | 0.002151 | 0.700708 | 0.005247 |
| Referral for dietary advice | 9 | 1 | same | 1 | same |
| Nutrition (must) | 8 | 1 | same | 1 | same |
| Norton Documentation - 3 times a week | 14 | -0.0769 | 0.7938 | -0.076923 | 0.793805 |

In terms of similarity in scores, we note that apart from the metric "skin state estimation - 3 times a week" and "Norton documentation - 3 times a week" all of the results were identical or highly significant, with most of the metrics having exactly the same scores. After investigating, we found that the reason for the difference in both cases was the incorrect date range selection of the nurse in the interface, so that the score she saw was not the correct score. Another difference was that instead of a score of 98 given by the system, the nurse gave a score of 100, probably because she saw a very high indicator and did not pay attention to the exact value.

These results demonstrate that by using the system after a short training, a senior nurse is able to search and retrieve accurately the score for a desired quality indicator for each patient.

4.5.4 Manual Quality Assessment Performance time when Assisted by the System

Figure 9 shows the time taken by the nurse to manually score all of the quality metrics for each patient. Note that there is a continuous improvement in the time taken to score all the quality metrics throughout patients. The mean time was 634.29 ± 303.87 seconds to score approximately 12 quality metrics per patient on average.

It should be noted that, even in a largely automated system, there is a significant minimum time for manual assignment of quality scores, since for each metric the nurse is required to perform a number of initial actions such as selecting dates, scrolling through the score tree to select another indicator, etc. Figure 9 shows the learning curve of the tool, and the clearly decreasing trend in QA performance time.

Note also that already by the 14th patient, the nurse's speed in providing quality scores for *all* the QA indicators, with the support of the system, was three times faster, compared to her mean speed when not using the system (360 seconds, compared to a mean of 1226 seconds), as depicted in Figure 8.
.

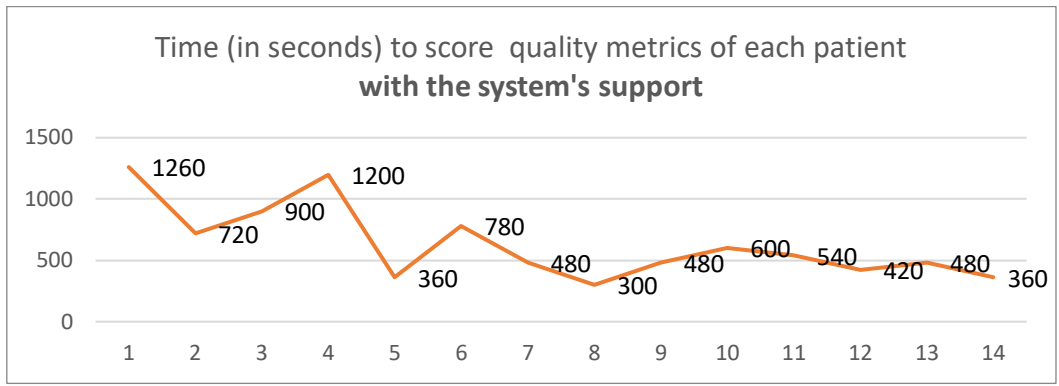

**Figure 9.** Time taken by the nurse to score all the quality metrics of each patient with the system's support.

4.5.5 Intermediate Conclusions from the Results

There were no significant differences ($P < 0.05$) across the values of most quality measures given by the nurse manually compared to the values given by the system automatically. That is, in practice, at least in the pressure ulcers GL case, the system appears to provide accurately and instantaneously quality assessment scores that are very similar to those of a highly experienced senior geriatric nurse.

There were also no significant differences ($P < 0.05$) between the values of most quality measures when the nurse used the system manually to retrieve specific QA scores for specific patients, and the scores retrieved by an experienced knowledge engineer using the same system; i.e., by using the system after a very short training period, a senior nurse is able to find the correct values of the quality measure for each patient.

Using the system's assistance significantly reduced the average assessment time it took for the nurse to score quality measures for each patient across all quality measures, reducing the time. By the 14th patient, the duration of the nurse's semi-manual assessment decreased to a third of her mean manual assessment time without the system's assistance.

5. Summary and Discussion

In this study, we designed, developed, and assessed a general system for quality control of medical care, based on the representation of existent medical guidelines. The system was evaluated using the protocol for the treatment of pressure ulcers.

The system includes a temporal fuzzy logic algorithm and considers for each action its type, its weight within the overall treatment protocol, and its partial fuzzy logic membership function. In addition, a friendly Web-based BI dashboard interface was developed, enabling users to display the performance of care providers from different aspects, such as for various treatment stages, several hospital wards, different patient groups, etc.

The system enables the medical staff to *identify gaps* in the application of a particular treatment protocol, compare the QA scores between different stages of the protocol (e.g. prevention versus follow-up), compare different wards and even different time windows, thus enabling an early detection of patient-management problems.

Using QA systems regularly is likely to help in increasing consistency and *reducing the variance in care*, or the so-called *noise* in medical management of patients [13]. As Kahneman et al. [2022] note, adherence to best-practice GLs certainly reduces noise.

In addition, the system allows the paramedical staff to closely monitor their own patients. For example, they may use the system to track protocol performance metrics in real time, and continuously through BI dashboard interface, and quickly identify patients for whom performance metrics are not high. We believe, given the vast literature on the benefits of providing feedback (e.g., as emphasized in the well-known book "*Nudge*" by Thaler and Sunstein), that such *an immediate QA feedback might well lead to an improvement in behavior*.

The system may also assist in the transferring of therapeutic responsibility to the nurses, when performing protocol-based actions, and thus *enhance their empowerment*, which, in addition to benefiting nurses, is also likely to free up time for physicians to address other, perhaps more clinically challenging problems.

Furthermore, when a new protocol becomes available online, the system has the potential ability, by providing direct and immediate feedback to the nursing staff, after each treatment, and after a certain minimal time period (day, week), to help all nurses adjust the implementation of the protocol to the standard that is expected. Thus, the system has the potential to help *preserve medical knowledge*, apply it at the treatment point in a relatively short time, and distribute it among nurses.

Even our limited evaluation has shown that the system saves time, whether used in automated or in semi-automated fashion. As the use of the system continues, and users become more experienced in operating the system for all its advanced functions, we expect that time savings will increase, while maintaining the same performance.

In a future research study, we plan to complete the experiment and perform a prospective experiment, using also the Picard GL-application engine, in a manner similar to our EU MobiGuide study and to our preeclampsia study. We will assess whether using the decision support system in real time increases the nurse's level of completeness and correctness, with respect to the application of evidence-based GLs, compared to a control group that is not provided with decision support. Thus, we shall move into the *quality-assurance* area, which we believe might be the best way to increase *quality-assessment* scores – namely, by preventing errors to begin with.

## Acknowledgements

This research was partially funded by the Israeli National Institute for Health Policy Research.

We would like to acknowledge the Herzfeld staff members who had assisted us in extracting the data, and the BGU 4[th] year Software and Information Systems Engineering students who had assisted us in implementing the quality-assessment system as part of their senior thesis.

**Appendix A - A description of the weights of each step and action acquired from the domain experts.**

| Stage | Stage weight | Component | Weight | Component | Weight | Component | Weight |
|---|---|---|---|---|---|---|---|
| Admission | 26% | Albumin | 12% | | | | |
| | | Diet | 22% | Diet nurse | 0.5 | | |
| | | | | Diet nutritionist | 0.5 | | |
| | | Pain | 12% | | | | |
| | | Norton | 27% | | | | |
| | | Skin | 27% | | | | |
| Total | | 100% | | | | | |
| Follow-up | 22% | Pain once a day | 30% | Performance | 0.5 | Frequency | 0.5 |
| | | | | | | Order | 0.5 |
| | | | | Command | 0.5 | | |
| | | Skin 3 times a week | 35% | | | | |
| | | Norton 3 times a week | 35% | | | | |
| Total | | 100% | | | | | |
| Prevention and follow-up | 26% | Pain once a day (big estimation) | TBD | Performance | 0.5 | Frequency | 0.5 |
| | | | | | | Order | 0.5 |
| | | | | Command | 0.5 | | |
| | | Skin 3 times a week (big estimation) | 10% | | | | |
| | | Norton 3 times a week (big estimation) | 10% | | | | |
| | | lubrication (frequency according to table). Olive oil, silicone, baby cream | 10% | Performance | 0.5 | Frequency | 0.5 |
| | | | | | | Order | 0.5 |
| | | | | Command | 0.5 | | |
| | | Leg lifting, Norton below 14 | TBD | Performance | 0.5 | Frequency | 0.5 |
| | | | | | | Order | 0.5 |
| | | | | Command | 0.5 | | |
| | | Skin once a day | 10% | Performance | 0.5 | Frequency | 0.5 |
| | | | | | | Order | 0.5 |
| | | | | Command | 0.5 | | |

| Category | % | Task | Weight | Sub-task | Value | Detail | Value |
|---|---|---|---|---|---|---|---|
| | | Instruction documentation once | 5% | | | | |
| | | Change of position 6/9 times a day | 20% | Command | 0.5 | | |
| | | | | Performance | 0.5 | Frequency | 0.5 |
| | | | | | | Order | 0.5 |
| | | Lay at 10 degrees once a day | 15% | Command | 0.5 | | |
| | | | | Performance | 0.5 | Frequency | 0.5 |
| | | | | | | Order | 0.5 |
| | | Nutrionist consultation | 10% | Nutritionist consultation | 0.5 | | |
| | | | | Nurse consultation | 0.5 | Reference | 0.5 |
| | | | | | | Consultation | 0.5 |
| | | Pressure dissipation | 10% | | | | |
| Total | | | 100% | | | | |
| Prevention and treatment follow-up (not including dressings) | 21% | Pain once a day (big estimation) | TBD | Performance | 0.5 | Frequency | 0.5 |
| | | | | | | Order | 0.5 |
| | | | | Command | 0.5 | | |
| | | Skin 3 times a week (big estimation) | 15% | | | | |
| | | Norton 3 times a week (big estimation) | TBD | | | | |
| | | Skin once a day | 15% | Performance | 0.5 | Frequency | 0.5 |
| | | | | | | Order | 0.5 |
| | | | | Command | 0.5 | | |
| | | Change of position 6/9 times a day | TBD | Command | 0.5 | | |
| | | | | Performance | 0.5 | Frequency | 0.5 |
| | | | | | | Order | 0.5 |
| | | Lay at 10 degrees once a day | TBD | Command | 0.5 | | |
| | | | | Performance | 0.5 | Frequency | 0.5 |
| | | | | | | Order | 0.5 |
| | | Pressure dissipation | TBD | | | | |
| | | Leg lifting, Norton below 14 | TBD | Performance | 0.5 | | |
| | | | | Command | 0.5 | | |
| | | Report to doctor | 15% | | | | |

| | | | | | |
|---|---|---|---|---|---|
| | | Document diagnosis | 5% | | |
| | | Visit with doctor and nurse once a week | 15% | Signature | ?? |
| | | | | Sore doctor visit | ?? |
| | | | | Sore nurse visit | ?? |
| | | Assess the integrity of the skin once a week (wound assessment) | 15% | | |
| Total | | | 80% | | |
| Dressings | 5% | Olive oil dressing | 34% | Frequency | 0.5 |
| | | | | Order | 0.5 |
| | | Foam and anti-odor dressings | 33% | Frequency | 0.5 |
| | | | | Order | 0.5 |
| | | Dressing Nacl0.9% | 33% | Frequency | 0.5 |
| | | | | Order | 0.5 |

# Appendix B - The SUS Questionnaire as filled up by the head nurse, Ms. Maya Salimanova.

**System Usability Scale (SUS)**

This is a standard questionnaire that measures the overall usability of a system. Please select the answer that best expresses how you feel about each statement after using the website today.

|    | | Strongly Disagree | Somewhat Disagree | Neutral | Somewhat Agree | Strongly Agree |
|----|---|---|---|---|---|---|
| 1. | I think I would like to use this tool frequently. | □ | □ | □ | □ | X |
| 2. | I found the tool unnecessarily complex. | □ | □ | □ | X | □ |
| 3. | I thought the tool was easy to use. | □ | □ | □ | □ | X |
| 4. | I think that I would need the support of a technical person to be able to use this system. | X | □ | □ | □ | □ |
| 5. | I found the various functions in this tool were well integrated. | □ | □ | □ | □ | X |
| 6. | I thought there was too much inconsistency in this tool. | X | □ | □ | □ | □ |
| 7. | I would imagine that most people would learn to use this tool very quickly. | □ | □ | □ | X | □ |
| 8. | I found the tool very cumbersome to use. | X | □ | □ | □ | □ |
| 9. | I felt very confident using the tool. | □ | □ | □ | □ | X |
| 10. | I needed to learn a lot of things before I could get going with this tool. | X | □ | □ | □ | □ |